\documentclass[runningheads]{llncs}
\usepackage{graphicx}
\usepackage{fancyhdr}
\usepackage{float}
\usepackage{hyperref}
\usepackage{todonotes}
\usepackage{caption}
\usepackage{xcolor}

\begin{document}

\title{Facial Landmark Visualization and Emotion Recognition Through Neural Networks}


\author{Israel Juárez-Jiménez, Tifanny Guadalupe Martínez-Paredes, Jesús García-Ramírez \and Daniel Sánchez-Ruiz \and Eric Ramos-Aguilar}

\institute{Unidad Profesional Interdisciplinaria de Ingeniería Campus Tlaxcala IPN, Tlaxcala, México \\
\email{ijuarezj2200@alumno.ipn.mx, tmartinezp2200@alumno.ipn.mx, jegarciara@ipn.mx, dsanchezro@ipn.mx, eramosa@ipn.mx}}


\maketitle

\begin{abstract}
Emotion recognition from facial images is a crucial task in human-computer interaction, enabling machines to learn human emotions through facial expressions. Previous studies have shown that facial images can be used to train deep learning models; however, most of these studies do not include a through dataset analysis. Visualizing facial landmarks can be challenging when extracting meaningful dataset insights; to address this issue, we propose facial landmark box plots, a visualization technique designed to identify outliers in facial datasets. Additionally, we compare two sets of facial landmark features: (i) the landmarks' absolute positions and (ii) their displacements from a neutral expression to the peak of an emotional expression. Our results indicate that a neural network achieves better performance than a random forest classifier.
\end{abstract}







\section{Introduction}

Emotion recognition through facial expressions is a key area of research in human-computer interaction and artificial intelligence, as facial expressions are one of the primaries means of non-verbal communication. Different methods have addressed this challenge, including traditional models such as random forests~\cite{munasinghe2018facial} and modern approaches based on deep neural networks~\cite{li2020deep}.

Some of the related works are based on extracting geometric features from facial landmarks~\cite{munasinghe2018facial}, while others have used convolutional neural network architectures to learn representations directly from images~\cite{li2020deep}, and also, a combination of both~\cite{liu2018conditional}. Statistical methods have previously been applied to identify patterns in facial points, focusing on discriminating between emotions such as happiness, sadness, or surprise~\cite{luo2015locating}. However, these techniques often face limitations, such as high data dimensionality or the need for large amounts of labeled data.

Despite advances, one of the main challenges is efficiency in data processing and analysis. Traditional methods, such as decision trees, are computationally less expensive, but present limitations in accuracy and generalization capacity. Deep neural network models address these issues, but they require more training time and significant computational resources.

This work proposes an approach that combines statistical methods with neural networks to improve emotion recognition in facial images. Unlike other studies, this method uses facial landmark normalization based on nose centering and dimensionality reduction through quartile analysis. To visualize the facial dataset, we propose facial landmark box plots in order to determine atypical data. 

Additionally, an improved neural network is implemented with Batch Normalization techniques and GELU activations, optimized to handle the extracted features efficiently. The proposed approach includes a comparison of two models: a decision tree and a neural network. Experimental results show that, while the decision tree achieved an accuracy of 80\%, the optimized neural network performed exceptionally well with an accuracy of 98.48\%. 



This paper is organized as follows: in Section~\ref{sec:related work} we present the related work; section~\ref{sec:methodology} introduce the proposed methodology; section~\ref{sec:results} shows the experimental results for the data visualization and classification stage; finally, section~\ref{sec:conclusions} presents the conclusions and future work.

\section{Related Works}
\label{sec:related work}

Several works have addressed facial emotion recognition using methods based on landmarks and neural networks~\cite{di2023randomized,tautkute2018know}. Among them, some studies have employed traditional statistical techniques to analyze the geometric features of the face~\cite{yao2021action}, while others have explored more advanced models that use deep neural networks to learn representations directly from images.

Facial landmark analysis has been widely used in previous research~\cite{haghpanah2022real}. For example, methods based on calculating distances between facial key points have proven effective in discriminating emotions such as happiness and sadness. However, these approaches often face limitations such as sensitivity to the initial position of landmarks and reliance on precise preprocessing to normalize the data.

In terms of neural networks, some works have implemented deep learning models that train convolutional networks to classify emotions directly from facial images. For instance, Picazo et al.~\cite{picazo2018redes} used manually labeled datasets to train networks for emotion recognition with high accuracy. However, these methods demand substantial labeled data and lengthy training times.

Other approaches have explored hybrid architectures that combine traditional geometric processing techniques with deep neural networks. For example,~\cite{pellejero2017analisis} proposed a two-step scheme in which a hidden representation is learned using convolutional layers, which is then transferred to a target task using a custom loss function. Despite promising results, this approach has high computational complexity, as it involves training multiple models sequentially.

In addition, initial efforts have been made to apply transfer learning strategies to emotion recognition~\cite{Xue_2021_ICCV}. Fine-tuning techniques have been experimented with to reuse convolutional layers from pre-trained models on different datasets~\cite{akhand2021facial}.

The work most similar to ours is~\cite{barbozareconocimiento}, who used a VGG neural network to find hidden representations that can be used as feature extractor to classify different emotions.  Unlike previous studies, our approach reduces the dimensionality of the data using quartile analysis and normalizes facial landmarks, which significantly improves the accuracy and efficiency of the final model.

\section{Proposed Methodology}
\label{sec:methodology}

\begin{figure}
    \centering
    \includegraphics[width=12cm]{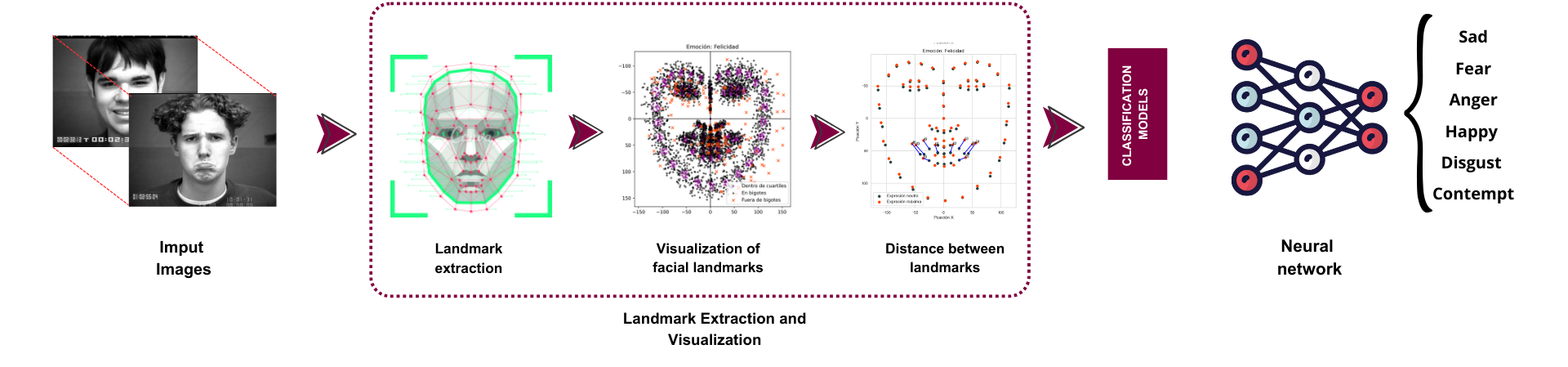}
    \caption{Proposed methodology for emotion recognition.}
    \label{fig:methodology}
\end{figure}

This section presents the proposed methodology for emotion recognition and data visualization. The proposed methodology can be seen in Figure~\ref{fig:methodology}. First, we perform a preprocessing stage in which we select the dataset. Next, we apply a deep learning model to extract facial landmarks from the images. In the following stage, we conduct a statistical analysis, introducing boxplots of the detected facial landmarks. Finally, we employ Random Forest and neural networks for emotion classification.

\subsection{Preprocessing stage}

\begin{figure}[htbp]
    \centering
    \includegraphics[width=0.9\linewidth]{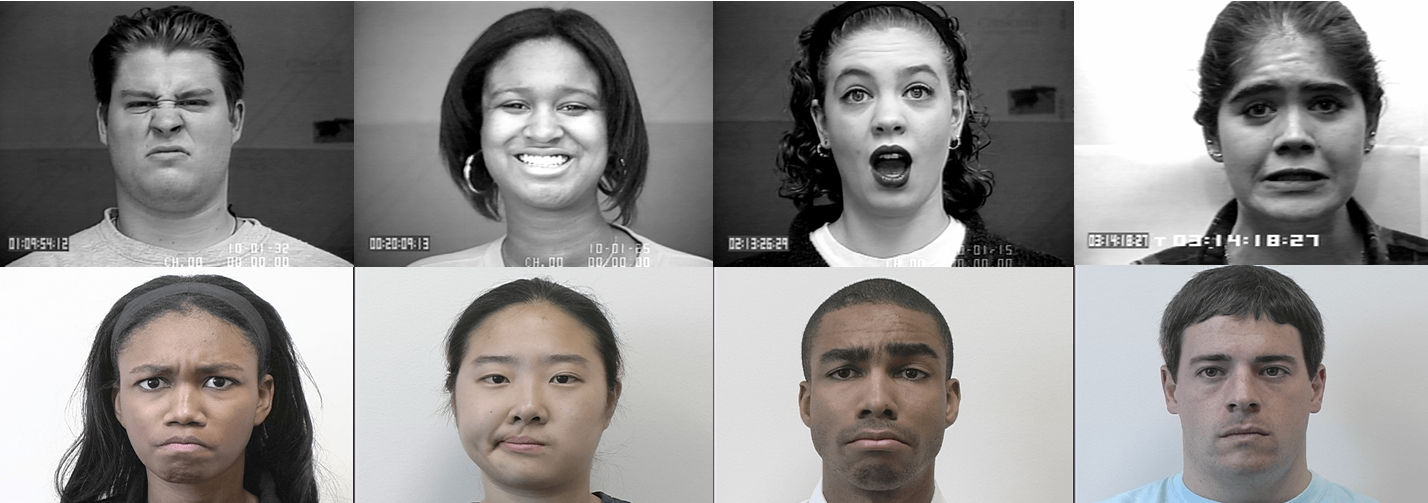}
    \caption{Examples of the CK+ dataset, this dataset is used in our experiments.}
    \label{fig:examples}
\end{figure}

The preprocessing stage is described in this section. For this study, we use the CK+ dataset, which contains facial images transitioning from neutral expressions to maximum emotions. The CK+ datasets includes 593 sequences that had a nominal emotion labeled based on the subject impression of 7 emotions: anger, disgust, contempt, fear, happy, sadness and surprise~\cite{lucas2010CK}. Some examples of the dataset can be seen in the Figure~\ref{fig:examples}.

The preprocessing consists of three main steps. First, we select two key images for each subject: the first frame, which represents a neutral expression, and the last frame, which captures the maximum emotional expression. Then, we perform a validation step to identify and remove incomplete entries or inconsistencies between emotion labels and images.

The process involved selecting the first frame (neutral expression) and the last frame (peak emotion) from each sequence in the CK+ dataset \cite{lucas2010CK}. The decision was based on the dataset's predefined labels, and no additional annotators were required as the labels were already validated. The validation step ensured consistency by removing sequences with missing frames or mismatched labels, resulting in a curated dataset ready for analysis.

After validation, the dataset consisted of paired images (neutral and peak emotion) for each of the 7 emotions: anger, disgust, fear, happiness, sadness, surprise, and contempt. The final dataset was balanced by addressing inconsistencies, ensuring each emotion had sufficient representation, though class imbalance remained (e.g., surprise was overrepresented, as shown in Fig.~\ref{fig:distribucion_emociones}). This step was critical for reliable feature extraction and model training.

At the end of this process, we obtain a structured and curated dataset, ready for facial landmark extraction and further analysis.

\subsection{Landmark detection}

Facial landmark detection and analysis involve several key steps to extract meaningful features from the images. First, the study employed Dlib's pre-trained model, which detects 68 facial landmarks using a histogram of oriented gradients (HOG) feature-based detector. This model was chosen for its robustness and accuracy in identifying key facial features. Once the face is detected, we apply a pre-trained model to identify 68 facial landmarks~\footnote{\href{https://huggingface.co/spaces/asdasdasdasd/Face-forgery-detection/resolve/ccfc24642e0210d4d885bc7b3dbc9a68ed948ad6/shape\_predictor\_68\_face\_landmarks.dat}{(Face forgery detection)}}. These landmarks correspond to key facial features, including the eyes, eyebrows, nose, and mouth.

To ensure consistency in the extracted features, we perform a normalization step. The landmarks are centered with respect to the midpoint between the eyes and scaled using Min-Max normalization, which eliminates variations due to differences in face position and size. Finally, for each subject, we compute the differences between the landmarks in the neutral and peak expression images. This process generates feature vectors that capture the facial deformations associated with different emotions.

\subsection{Statistical Analysis}

To enhance the quality of the extracted features, we perform a statistical analysis in two main stages. First, we identify outliers by computing quartiles and the interquartile range (IQR) for the landmark coordinates. Any values falling outside the limits defined by $Q1-1.5 \cdot IQR$ and $Q3+1.5 \cdot IQR$ are considered outliers and handled accordingly, either by removing them or adjusting their values to mitigate their impact on the model.

Next, we categorize the landmark points into three groups based on their statistical distribution. Points within the quartile range are considered normal and representative of the central distribution. Those located within the whiskers of the boxplot are near the quartile boundaries but are not classified as outliers. Finally, points falling outside the whiskers are identified as outliers and require special treatment.

Additionally, we generate various visualizations to better analyze and understand the data. Histograms and pie charts are used to illustrate the distribution of emotions within the dataset, ensuring a balanced representation of each class for classification purposes. Scatter plots and boxplots are also constructed to examine the distributions and variability of the distances between landmarks in neutral and peak emotion expressions.

This statistical analysis ensures the quality and representativeness of the features used in the classification models.

\subsection{Classification stage}

Two models were implemented and evaluated for emotion classification: one based on decision trees and the other using neural networks with different configurations. We select this algorithms because we want to compare an algorithm that is much interpretable as Random Forest and other that obtain an acceptable performance in classification tasks as Neural Networks.

The first model is a decision tree, an interpretable method that organizes the features into a binary hierarchical structure~\cite{breiman2001random}. In this model, successive splits of the data are made based on their importance, determined using criteria such as gini or entropy. Key hyperparameters were tuned, such as the maximum depth of the tree, the splitting criterion, and the minimum number of samples per leaf. This model served as an initial reference to assess the complexity of the problem and to establish a baseline for comparison with more advanced models.

Subsequently, two neural network models were implemented with different configurations. The first was a basic neural network, consisting of two fully connected layers: the first with 256 neurons and the second with 128 neurons, both using the ReLU activation function and regularized with 50\% Dropout, this also stabilize the learning curves. The output layer contained 7 neurons, corresponding to the emotional classes, and used the Softmax activation function. For training, the Adam optimizer, cross entropy loss function, and a training schedule of 50 epochs were employed.

To improve performance, an optimized neural network was designed, incorporating more advanced architectures. This network consisted of three fully connected layers: the first with 512 neurons and GELU activation, followed by a second layer with 256 neurons and a residual connection to enhance gradient propagation. The third layer had 128 neurons, maintaining the GELU activation across all layers. Additionally, Batch Normalization was applied to stabilize training, and 30\% Dropout was used to prevent overfitting. The Adam optimizer with an initial learning rate of 0.001 and a learning rate scheduler were used to improve model convergence.

\section{Experimental Results}
\label{sec:results}

In this section we present the experimental results for the data analysis and then we present the results of the training stage.

\subsection{Data visualization}

\begin{figure}[htbp]
    \centering
    \includegraphics[width=0.8\linewidth]{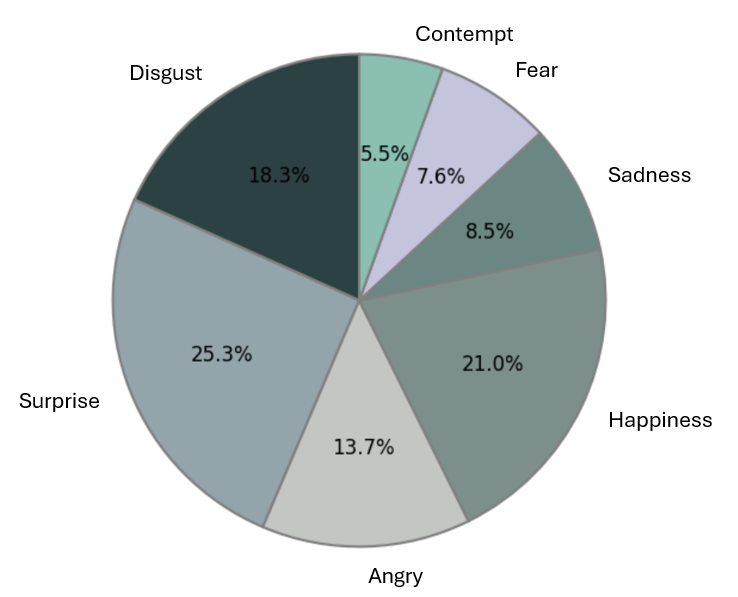}
    \caption{Distribution of the classes in the CK+ dataset, disgust, happiness and surprise are the classes with more instances.}
    \label{fig:distribucion_emociones}
\end{figure}

In the Fig.~\ref{fig:distribucion_emociones} we show the distribution of the data of the used dataset, in this case CK+~\cite{lucas2010CK}, where seven labeled emotions are contained. We can see that there is a clear imbalance of the classes, surprise is the class with more data, while contempt and fear are the classes with low instances, respectively. The imbalance can bias the model to the more representative classes.

Also, we introduce facial landmark boxplot, where the distribution of the data can be seen in a single figure instead of use one box plot for each landmark. The main idea is to plot the entire points of the landmarks taking into account the points outside the IQR in a different mark, in order to visualize how many of them are in the data distribution.

\begin{figure*}[htbp]
    \centering
    \noindent
    \includegraphics[width=0.9\linewidth]{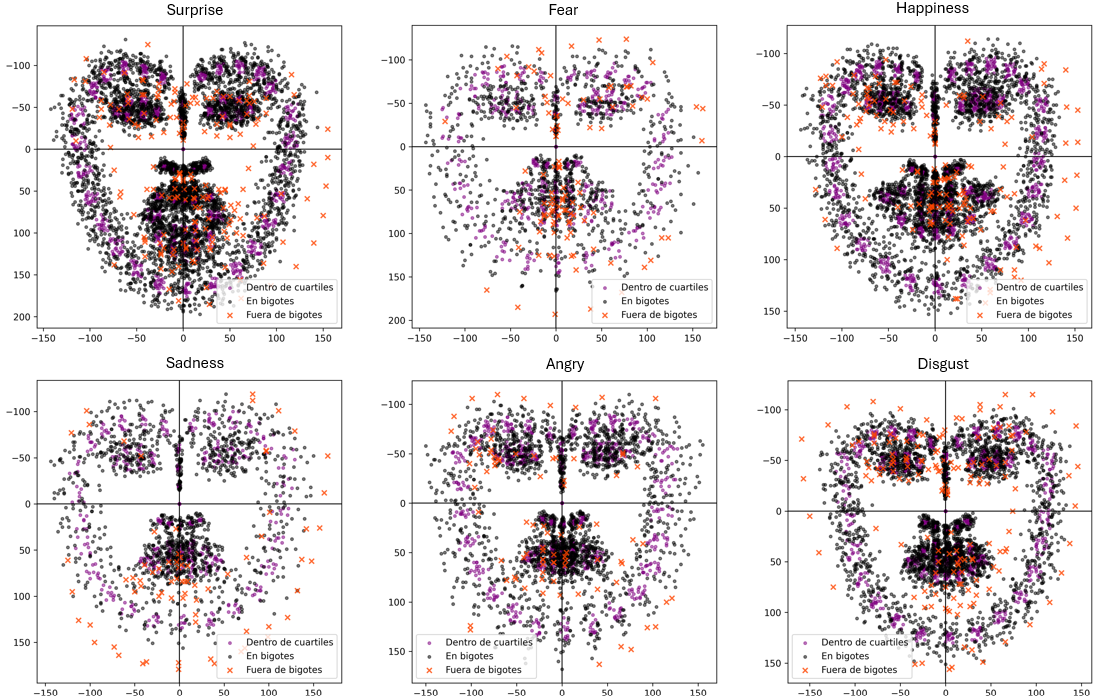}
    \caption{Examples of facial landmarks boxplots for Surprise, Fear, Happiness, Sadness, Angry and Disgust emotions. We can see low number of outliers in the images of the three classes, the main advantages is that we can see the entire landmark boxplot in a figure, instead of see one boxplot for each landmark.}
    \label{fig:boxplot}
\end{figure*}

Figure \ref{fig:boxplot} illustrates the distribution of emotions in the CK+ dataset, revealing a significant imbalance. Surprise is the most represented emotion, accounting for 25.3\% of the dataset, while Contempt and Fear are the least frequent, with 5.5\% and 7.6\%, respectively. This imbalance could bias the model toward the more common emotions, such as surprise and happiness, while making it more challenging to recognize the less frequent ones. Therefore, it is crucial to consider techniques to address this imbalance and ensure fair classification across all emotions.

Table \ref{tab:emotion_distribution} presents the distribution of facial landmarks within the whiskers and the percentage of outliers for each emotion. Contempt exhibits the highest proportion of points within the whiskers (68.71\%), indicating a more concentrated and uniform distribution. In contrast, fear shows the highest percentage of outliers (6.12\%), suggesting greater variability in its expression. Surprise, with the largest number of total points (5644), maintains a high proportion of points within the whiskers (66.85\%) and a low outlier rate (3.19\%), reflecting its strong representation in the dataset. Emotions such as anger, disgust, and sadness display similar distributions, with approximately 64-66\% of points inside the whiskers and around 3-4\% classified as outliers. The results highlight the potential influence of data imbalance on model training, emphasizing the need for techniques to mitigate biases and ensure an equitable classification across all emotions.

\begin{table*}[htbp]
    \centering
    \caption{Distribution of points within whiskers and outliers for each emotion.}
    \label{tab:emotion_distribution}
    \begin{tabular}{|p{2cm}|p{1.6cm}|p{1.6cm}|p{6.5cm}|}
        \hline
        \textbf{Emotion} & \textbf{\% in Whiskers} & \textbf{\% Outliers} & \textbf{Observations} \\
        \hline
        Anger      & 64.51\%  & 2.78\%  & Expected distribution, low variability. \\
        Contempt   & 68.71\%  & 3.76\%  & More concentrated and uniform distribution. \\
        Disgust    & 64.36\%  & 3.91\%  & Variability within quartiles: 31.73\%. \\
        Fear       & 58.71\%  & 6.12\%  & Highest variability, more outliers. \\
        Happiness  & 62.47\%  & 3.39\%  & Proportion within quartiles: 34.14\%. \\
        Sadness    & 65.70\%  & 3.83\%  & Moderate proportion within quartiles: 30.46\%. \\
        Surprise   & 66.85\%  & 3.19\%  & Largest total number of points (5644), well-represented. \\
        \hline
    \end{tabular}
\end{table*}

\subsection{Classification stage}

In our experiments we use two popular algorithms, neural networks and random forests because these work well with similar data to ours. Nevertheless, random forests shows poor performance compared with neural networks. We perform the experiments using a 4-fold cross validation, that and we present the learning curves for the best model. In order to compare the classifiers we present the mean of the for experiments in the next paragraphs.

The Random Forest model achieved an accuracy of 80\% . While this performance is moderate, it suggests that the model effectively captures relevant patterns in the data. However, the complexity of the extracted features may have limited further improvements. In comparison, the Decision Tree model, despite its interpretability, exhibited suboptimal results, reinforcing the need for more sophisticated techniques to enhance classification performance. These findings highlight the importance of balancing model complexity and interpretability to achieve robust emotion recognition.

In the case of neural networks, we apply the configuration described in the previous section. In our experiments we observe that the loss on the test set stabilizes after several epochs and exhibits slight variability, which is expected in generalization processes. The minimal gap between training and test loss in the final epochs indicates that the model has achieved good generalization without significant overfitting. This suggests that the model can correctly classify new samples without being overly influenced by specific patterns in the training set.

\begin{figure}[htbp]
    \centering
    \includegraphics[width=0.7\textwidth]{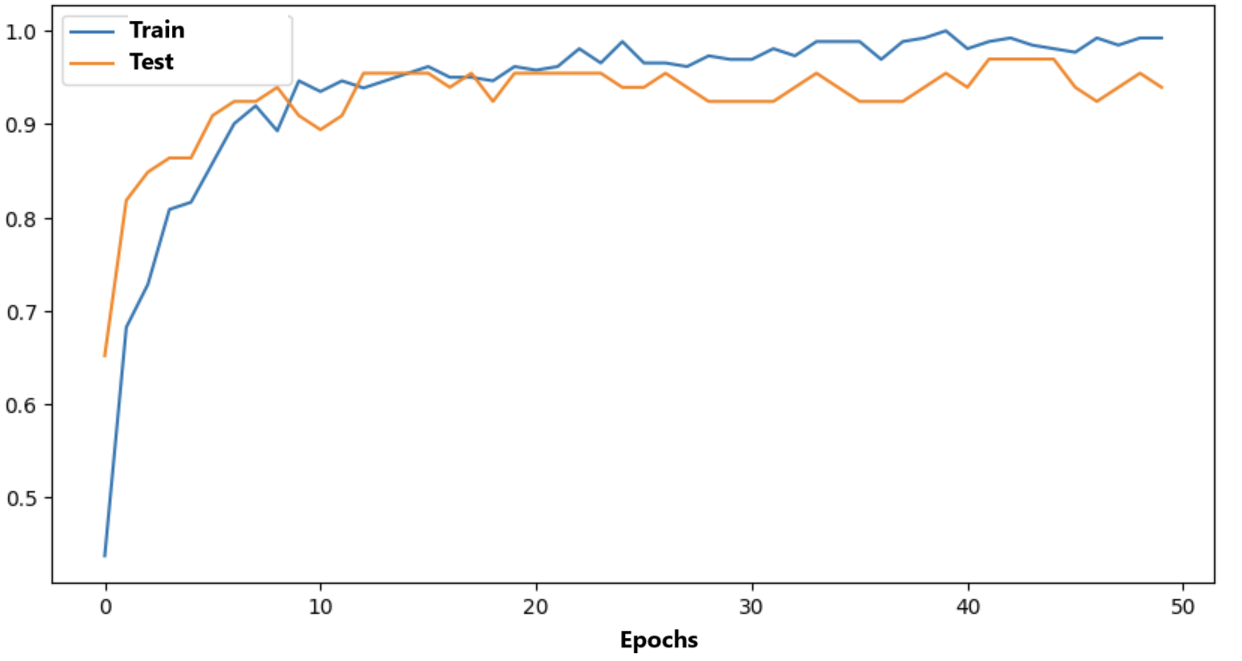} 
    \caption{Accuracy learning rate for the neural network that use ReLU activation function.}
    \label{fig:precision_rn_basica}
\end{figure}

\begin{figure}[htbp]
    \centering
    \includegraphics[width=0.7\textwidth]{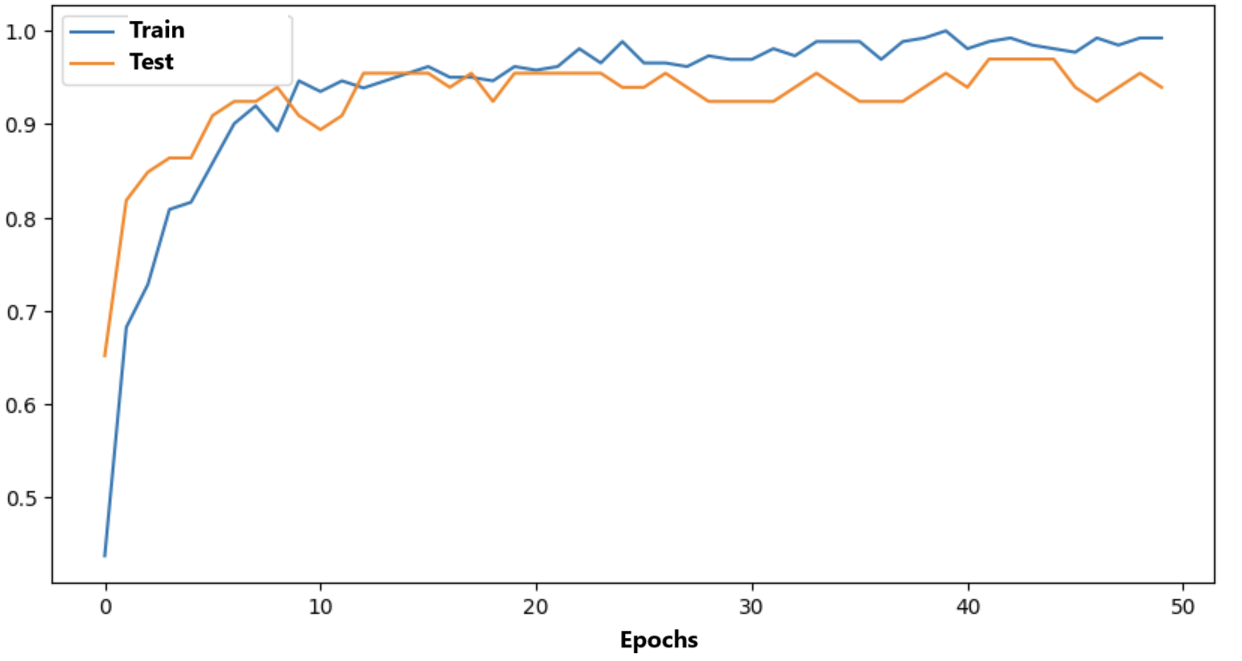} 
    \caption{Loss for the neural network that use ReLU activation function.}
    \label{fig:loss_ReLU}
\end{figure}

Figure~\ref{fig:precision_rn_basica} shows the training and test accuracy over epochs with the neural network that use ReLU activation function (Also, it can be seen the loss of the training in Figure~\ref{fig:loss_ReLU}). Similar to the loss behavior, the training accuracy improves rapidly in the initial epochs, reaching values close to 100\% by the end of training. This indicates that the model has effectively learned the patterns present in the training set.

In the test set, accuracy increases quickly in the early epochs and stabilizes around 96.97\%. The close proximity between training and test accuracy curves suggests that the model generalizes well, achieving high accuracy on unseen data without overfitting. This demonstrates the model’s robustness and its ability to accurately classify emotions in facial images.

\begin{figure}[htbp]
    \centering
    \includegraphics[width=0.7\textwidth]{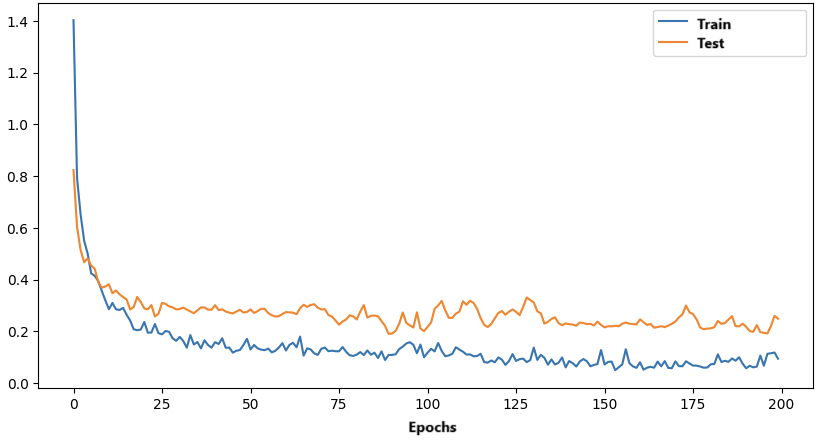} 
    \caption{Loss learning curve for the neural network that use GeLU activation function.}
    \label{fig:perdida_rn_opt}
\end{figure}

Figure \ref{fig:perdida_rn_opt} illustrates the evolution of loss during training and evaluation with the GeLU activation function. Initially, the loss is high but rapidly decreases in the early epochs, indicating that the model quickly learns the fundamental patterns of emotions. As training progresses, the training loss continues to decrease, approaching zero, suggesting a good fit to the data.

\begin{figure}[htbp]
    \centering
    \includegraphics[width=0.7\textwidth]{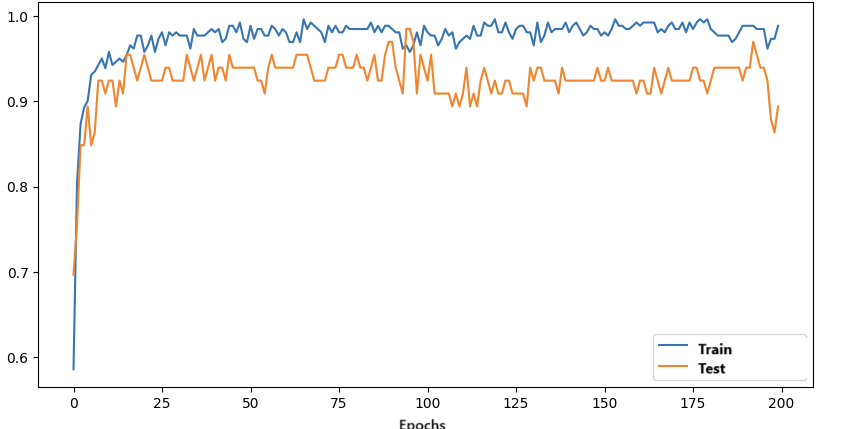} 
    \caption{Accuracy learning curve for the neural network that use GeLU activation function.}
    \label{fig:precision_rn_opt}
\end{figure}

Finally, Figure \ref{fig:precision_rn_opt} presents the accuracy throughout training and testing. The model achieves near 100\% accuracy on the training set and 98.48\% on the test set. This performance highlights the model’s ability to correctly classify emotions, even on unseen data, demonstrating its robustness and effectiveness in the given task.

\section{Conclusions and Future Work}
\label{sec:conclusions}

This work presented an integrated approach for facial emotion recognition, using the CK+ dataset for feature extraction and classification. The proposed method included key steps such as data preprocessing, facial landmark normalization, statistical analysis for feature enhancement, and classification using models of varying complexity.

The evaluation of different models provided valuable insights into their effectiveness. The Decision Tree classifier, while interpretable and computationally efficient, achieved a moderate accuracy of 80\%, struggling with complex feature representations. The Basic Neural Network significantly outperformed it, reaching over 96\% accuracy, making it a suitable choice for real-time applications due to its fast training and inference. The Optimized Neural Network emerged as the most effective model, achieving 98.48\% accuracy, demonstrating strong generalization capabilities through advanced architecture, batch normalization, residual connections, and hyperparameter tuning. However, this higher accuracy came at the cost of increased training time.

The comparison highlights the trade-off between simplicity, speed, and accuracy when selecting a model. Simpler models are preferable for fast tasks, whereas more complex architectures offer higher accuracy but require greater computational resources.

Future work could explore transfer learning with pretrained models to improve performance in real-world scenarios, where lighting, pose variations, and demographic diversity impact facial expressions. Additionally, data augmentation techniques could enhance generalization by addressing the dataset’s limited variability. Finally, developing lightweight models optimized for low-resource devices could enable practical applications in real-time monitoring and mobile systems.

\bibliographystyle{splncs04}
\bibliography{bib}

\end{document}